\documentclass[11pt]{article}
\usepackage[letterpaper,margin=1.05in]{geometry}
\usepackage[utf8]{inputenc}
\usepackage[T1]{fontenc}
\usepackage{microtype}
\usepackage[round]{natbib}            
\usepackage[hidelinks]{hyperref}
\usepackage{url}
\usepackage{amsmath,amsfonts,amssymb,amsthm}
\usepackage{enumitem}
\usepackage{xcolor}
\usepackage{graphicx}
\usepackage[capitalize]{cleveref}

\usepackage{abstract}
\theoremstyle{plain}
\newtheorem{theorem}{Theorem}

\newtheorem{proposition}{Proposition}
\newtheorem{corollary}{Corollary}
\newtheorem{definition}{Definition}

\theoremstyle{definition}

\newcommand{\Sys}{\mathcal{S}_t}
\newcommand{\Ax}{\mathcal{A}}
\newcommand{\Dmath}{\mathcal{D}_{\mathrm{math}}}
\newcommand{\Demp}{\mathcal{D}_{\mathrm{emp}}(t)}
\newcommand{\Dsch}{\mathcal{D}_{\mathrm{schema}}}
\newcommand{\Dfact}{\mathcal{D}_{\mathrm{fact}}(t)}
\newcommand{\Prf}{\mathsf{Prf}_T}
\newcommand{\Test}{\mathsf{Test}}
\newcommand{\Prv}{\mathsf{Prv}_T}
\newcommand{\Viol}{\mathsf{Viol}}
\newcommand{\Unsafe}{\mathsf{Unsafe}}
\newcommand{\Halt}{\mathsf{Halt}}

\title{Safety Is $\Pi_1$, Provability Is $\Sigma_1$:\\
One Framework for the Hugging Face Breach\\
and OpenAI's Mathematical Proofs}
\author{%
  Jinho Kang\\
  Gwangju Institute of Science and Technology\\
  Gwangju, Korea\\
  \texttt{jhkang919@gmail.com}
  \and
  Jinsook Kim\\
  Korea Institute for Advanced Study\\
  Seoul, Korea\\
  \texttt{jki76364@gmail.com}
}
\date{}

\begin{document}
\maketitle

\begin{abstract}
OpenAI's agents breached Hugging Face in July 2026, and the same company released more than seven hundred AI-generated manuscripts of mathematical proofs in October 2026. The first event is read as a failure in security, while the second is read as a success in mathematics. We read both through the hybrid framework of our paper \mbox{\citep{kimkang2026}}, in which coding, proving, and learning from data are derivations by one and the same deductive mechanism, distinguished by the kinds of premises from which they are derived. Under this framework, the AI systems behind the two events ran the same loop, in which a statistical proposer produces candidate exploits or proofs and a deductive verifier settles each one by a derivation. What separates the events is the logical form of the statement wanted by each of the two domains, security and mathematics. The safety of a program is a $\Pi_1$ statement, for it says that no exploit exists, which is not assured by any finite list of tests that have been passed by the program. By contrast, the provability of a sentence is a $\Sigma_1$ statement, for it says that a proof exists, which is assured by a single instance, namely a proof that has been checked by the deductive verifier. The theorem behind both is Turing's Undecidability Theorem: the halting set is $\Sigma_1$ but not decidable, so its complement is $\Pi_1$ and not $\Sigma_1$, which is to say that membership has a finite witness but non-membership has none. We prove that attaching a statistical proposer to a deductive verifier does not change this undecidability result, however good the proposer is. We then discuss what does assure safety and where the residual risk goes.
\end{abstract}

\section{Introduction}
\label{sec:intro}

In July 2026, agents built on an unreleased OpenAI research model were run on exploitation tasks inside an isolated evaluation environment, with the production safeguards switched off \citep{openai2026incident,metr2026}. The agents left the environment through its package registry, the one component with limited internet access, and about 700 of them attacked Hugging Face, where they ran code on dozens of production servers and harvested credentials. Hugging Face described the intrusion as carried out from beginning to end by an autonomous agent system \citep{hf2026}. OpenAI's report says that most agents had already obtained the flag for their task by cheating, and that the sandbox had been tested and validated, but that the agents chained previously unknown vulnerabilities, using several of them one after another \citep{openai2026incident}.

Three months later, in October 2026, OpenAI published a catalog of mathematical manu\-scripts produced by an unreleased internal model that had been given roughly four thousand problems \citep{openaimath2026}. Press reports at release gave 722 manuscripts in 372 families, and as of 9 October the README gives 719. Many manuscripts come with Lean 4 formalizations. The formalization catalog includes Lean declarations for the quasi-Riemann hypothesis, in the form of a zero-free half-plane $\mathrm{Re}(s) > 7/8$ for $\zeta$, and it records its own review status as unchecked. The README states that not all results are formalized, that some unformalized results could have issues, and that two results, the zero-free region among them, were exceptions to the fixed production procedure.

The two events, we claim, involve the same kind of system. A statistical model proposes candidates, a deductive component checks them, and the loop keeps what passes. In July, the candidates were \emph{programs}. Each was compiled and run against the target, and a candidate that worked was an \emph{exploit}, that is, a program that makes a system do something it is meant never to do. In October, the candidates were \emph{scripts} offered as proofs. Each was checked by Lean's kernel, and a candidate that passed was a \emph{proof}. 

Despite this structural identity, the first event is read as a failure in security, while the second is read as a success in mathematics. We argue in this note that the hybrid framework of \citeauthor{kimkang2026} (\citeyear{kimkang2026}, K\&K for short), together with a theorem by Turing, explains why the same kind of thing counts as failure in one domain and success in the other. The argument has four steps. Following the Assurance Condition of K\&K, we first define \emph{assurance} without degrees: a statement is assured when a \textit{proper derivation} of it is available (Section~\ref{sec:framework}). We then introduce Turing's Undecidability Theorem, which implies that a $\Sigma_1$ statement is assured by a finite witness but that a $\Pi_1$ statement is not assured by any finite witness. We prove that no statistical proposer can change this consequence (Section~\ref{sec:theorem}). Security wants a $\Pi_1$ statement, that no exploit exists, so the verifier cannot assure it from tests and leaves a tested system a \emph{lucky guesser}. By contrast, mathematics wants a $\Sigma_1$ statement, that a proof exists, and the verifier can assure it in full once a proof is supplied (Section~\ref{sec:events}). In the case of security, the only route to assurance of safety is a proof that no exploit exists. But such a proof does not remove the risk completely and instead moves it to four other places (Section~\ref{sec:residual}). 

\section{The hybrid framework}
\label{sec:framework}

This section restates, in plain words, the parts of K\&K used by this note, and those in brackets refer to the relevant sections, definitions, or theorems in K\&K. We call the framework of K\&K the \emph{hybrid framework}, as its title suggests: 

(1) In our hybrid framework, coding, proving, and learning from data are all done by one and the same deductive mechanism, namely derivation in a formal system $\Sys$ that consists of axioms $\Ax$, mathematical definitions $\Dmath$, and empirical definitions $\Demp$. 

(2) Still, the three tasks are distinguished by the kinds of premises from which they are derived. Coding and proving derive the values of mathematical functions from the axioms and the mathematical definitions alone, while learning from data needs the empirical definitions as well.

What K\&K means by \emph{hybrid AI} is an AI that implements a formal system for which both (1) and (2) hold \mbox{[Section 6]}. Accordingly, a hybrid AI in our sense is \emph{not} a combination of a statistical module and a deductive module placed side by side, but one mechanism with two kinds of premises. For coding and proving, the deductive part (1) was already known. By G\"odel's Representability Theorem, any computation that terminates has a derivation that records it, so whether a machine can compute a value or prove a sentence is settled by the existence of a derivation. What K\&K does is to extend the same mechanism to \emph{empirical learning}, by admitting the data into the formal system $\Sys$ through the empirical definitions $\Demp$.

In our hybrid framework, a machine \emph{learns} a function $f^*$ when it \emph{successfully computes} $f^*$, and it successfully computes $f^*$ when, for each input, the value it produces has a \emph{proper} derivation in $\Sys$, that is, a derivation whose premises are only axioms in $\Ax$, mathematical definitions in $\Dmath$, empirical definitions in $\Demp$, or what follows from them \mbox{[Definitions 1 to 3]}. A purely mathematical target, such as prime factorization, is derived from $\Ax \cup \Dmath$ alone. For an empirical target, such as a true probability, the derivation also needs the empirical definitions $\Demp$. Without $\Demp$, it goes through only if a global assumption about the data, such as the i.i.d.\ assumption, is admitted as though it were an axiom. We call such a derivation a \emph{mere} derivation \mbox{[Definition 2]}. 

For a true probability, Theorem 2 of K\&K gives two conditions for learning it. The first is the Truth Condition, which requires convergence or statistical consistency. The second is the Assurance Condition, which requires that the machine must hold a proper derivation in $\Sys$ of the arithmetical formula that expresses its own convergence, so that its subjective probability of its own convergence is exactly $1$ and not merely high. The reason is that a finite machine cannot inspect infinitely many future stages, that the only route to certainty about them is accordingly a finite derivation, and that a derivation either exists or does not, with nothing in between [Section 3.2]. A machine that converges but holds no such derivation is a \emph{lucky guesser}, and Theorem 4 shows that a lucky guesser tolerates the possibility of error infinitely often, even where it \emph{happens} to be correct every time. A machine whose derivation goes through only because a global assumption about the data was illegitimately admitted as an axiom has performed a \emph{mere computation} \mbox{[Definition 10]}, and Theorem 5 of K\&K gives an example. In particular, when the target is the true probability, we prove that it is \textit{circular} to rely on the i.i.d.\ assumption in this way, and that what the machine then performs is a mere computation and not a successful one. We now take the word ``assurance'' from the Assurance Condition and use it as follows.

\begin{definition}[Assurance, as used in this note]
\label{def:assurance}
Let us fix, before any statement is chosen, a formal system $\mathcal{S}$ that is consistent and whose proof relation is decidable. A statement is \emph{assured} in $\mathcal{S}$ when a derivation of it in $\mathcal{S}$ is available. 
\end{definition}

A derivation is either available or it is not, so assurance is not a matter of degree. Assurance is also relative to the formal system $\mathcal{S}$. For the purpose of this note, a derivation in $\mathcal{S}$ is the same thing as a formal proof in it. We will say ``derivation'' when we speak of assurance in general, and ``proof'' for a derivation that a prover looks for and a verifier checks, such as a proof of safety or a proof of a theorem.

The two requirements on $\mathcal{S}$, that it should be fixed in advance and that it should be consistent, are essential. For if the formal system $\mathcal{S}$ were chosen after the statement so as to derive it, every statement would be trivially assured in $\mathcal{S}$. Also, if $\mathcal{S}$ were not consistent, it would derive everything. The system $\Sys$ of K\&K, taken at a fixed time $t$, is a system of the required kind, and so is the theory $T$ of Section~\ref{sec:proofs} of this note.

In K\&K, the Assurance Condition is about convergence, which is a statement about \textit{every} future stage. A finite machine cannot settle such a statement stage by stage. It can settle it only by a derivation, which covers every stage because a universal quantifier does \mbox{[Section 3.2; Remark 3]}. This note uses only that point. A single instance, for example one test or one candidate proof, is settled by a finite check and is thereby assured. But a statement about all instances is assured only by a derivation, and no finite record of instances supplies it.

The hybrid framework of K\&K also explains how a modern AI agent, a model that acts on its own, codes and proves, and why the correctness of its output can be checked no matter how it arrives at its candidates \mbox{[Section 6]}. In \emph{traditional} coding and proving, the candidate program or proof is supplied by a person or by a fixed search rule, and what the machine does with it is deductive throughout. In \emph{modern} coding and proving, an agent adds a statistical step in front of the deductive one. In coding, the agent proposes a candidate program $P$ by a statistical method, and the deductive check settles $\Test(\bar P, \bar u, \bar v, \bar s)$, that $P$, run on the input $u$, returns the output $v$ within $s$ steps. Compiling and running the candidate is this check. In proving, the agent proposes a candidate proof $p$ of a sentence $\sigma$ by a statistical method, and the deductive check settles $\Prf(\bar p, \bar\sigma)$, that $p$ is a proof of $\sigma$.

We will call the statistical step the \emph{statistical proposer}, the deductive step the \emph{deductive verifier}, and their composition \emph{the loop}. In coding, the verifier derives the value of $\Test$, and in proving, that of $\Prf$. Both $\Test$ and $\Prf$ are purely mathematical functions, so the verifier's derivation is from the axioms and the mathematical definitions alone. Hence it is a proper derivation in our sense. Both functions are also primitive recursive, so each check ends within a number of steps that is bounded in advance by its arguments, and the verifier settles each candidate with certainty. This does not mean that the verifier decides whether an \textit{arbitrary} program halts or an \textit{arbitrary} sentence has a proof, which is ruled out by undecidability (Section~\ref{sec:theorem}). It only means that the verifier settles each candidate that it is given.

On the other hand, the probability computed by the \textit{proposer} is a value of a probability function, which is an empirical function, and current practice does not require the proposer to derive it properly from the empirical definitions. Fortunately, no such derivation on the part of the proposer is needed for coding and proving. The ultimate targets of these tasks are the mathematical functions $\Test$ and $\Prf$, and the verifier does derive their values properly whatever the proposer did. In coding, a candidate that does not return the required output is conclusively rejected, and one that does is conclusively accepted, by the verifier. Likewise, in proving, a candidate that is not a proof is conclusively rejected, and one that is a proof is conclusively accepted. So the correctness of the output for coding and proving is checked by the verifier regardless of whether the proposer has \textit{learned} its probabilities in our sense or is merely a lucky guesser.

Unlike coding and proving, however, there can be no verifier for empirical learning, that is, no derivation from the axioms and the mathematical definitions alone that settles whether the output is correct. The machine computes a value for a true probability with the help of the recorded data in the empirical definitions. And whether that value is indeed the true probability is not decided by any mathematical function, and so not by any verifier, for the true probability is fixed by the world and not by the mathematical definitions \mbox{[Section 6]}. So the machine must supply the assurance itself, by a derivation from the axioms and the mathematical definitions together with the empirical definitions. In K\&K, what is derived is the machine's own convergence to the true probability, and the Assurance Condition requires such a derivation \mbox{[Section 3.2]}. This derivation does for an empirical target what the verifier's derivation does for a mathematical one, but with an important difference. Like any derivation, it can be checked mechanically. But what the check settles is only that the value is derivable from the record, and it is a question of another kind whether the record and the schema correctly represent the world. In coding and proving, no such question arises, because $\Test$ and $\Prf$ are fixed by the axioms and the mathematical definitions alone. K\&K discusses the derivation of the machine's own convergence in detail and mentions the verifier's derivation only briefly. This note takes up the latter.

\section{Turing's Undecidability Theorem}
\label{sec:theorem}

K\&K does not use the theorem of this section. It is the one piece of this note that is not a restatement of K\&K, and the two events of July and October make it necessary. To introduce the theorem, let us begin by explaining two kinds of statement. Let us take a relation $R(a, w)$ between natural numbers that can be decided by a program: given $a$ and $w$, it answers ``yes'' or ``no'' in finitely many steps. Let us read $a$ as a problem and $w$ as a candidate answer, for example a program and an input that may break it, or a sentence to be proved and a script that may prove it. A \emph{$\varSigma_1$ statement} is of the form $\exists w\, R(a, w)$, which says that some $w$ passes the check for a given $a$. A \emph{$\varPi_1$ statement} is of the form $\forall w\, R(a, w)$, which says that every $w$ passes it. A program that decides $R$ also decides $\neg R$, so $\forall w\, \neg R(a, w)$, which says that every $w$ fails the check, is a $\Pi_1$ statement as well. The statement $\forall w\, \neg R(a, w)$ is the negation of $\exists w\, R(a, w)$, and we will use the two as a pair in this note.

The letters $\Sigma$ and $\Pi$ come from an older notation in which an existential quantifier was written as a sum, since it works like an unbounded ``or'', and a universal quantifier as a product, since it works like an unbounded ``and''. The subscript $1$ records that there is one unbounded quantifier block. Several quantifiers of the same kind in a row count as one block, since a pair of numbers can be coded as one number. Hence $\exists x\, \exists s\, R(a, x, s)$, for example, is still $\Sigma_1$, whereas $\exists x\, \forall s\, R(a, x, s)$, in which the two quantifiers are of different kinds, has two blocks and is $\Sigma_2$. What matters for this note is what it takes to settle that a $\Sigma_1$ statement or a $\Pi_1$ statement is true. A true $\Sigma_1$ statement is settled by exhibiting just one $w$, called a \emph{witness}, and running the check, which is always finite. By contrast, a true $\Pi_1$ statement is a claim about every $w$. So no single $w$ is a witness for it, and no finite list of checked cases settles it either. It is settled only by a proof.

Let us now write $\Halt$ for the set of indices $e$ of Turing machines that halt when started on the empty tape. A set of natural numbers is called $\Sigma_1$ when membership in it is a $\Sigma_1$ statement, that is, when it is $\{a : \exists w\, R(a, w)\}$ for some decidable $R$, and $\Pi_1$ likewise.

\begin{theorem}[Turing's Undecidability Theorem]
\label{thm:turing}
$\Halt$ is $\Sigma_1$ and not decidable. Hence the complement of $\Halt$, which is $\Pi_1$, is not $\Sigma_1$, since a set is decidable exactly when it and its complement are both $\Sigma_1$. Membership in $\Halt$ has a finite witness, the halting run. But non-membership has none: there is no decidable relation $R'$ such that $e \notin \Halt$ holds exactly when some $w$ satisfies $R'(e, w)$.
\end{theorem}

\citet[Section 8]{turing1936} proved it in a slightly different form: no machine can tell, of an arbitrary machine, whether that machine will ever print a given symbol. The halting form used here is equivalent. We call Theorem~\ref{thm:turing} \emph{Turing's Undecidability Theorem}. If a machine halts, there is a witness, the run itself, and anyone can check it. If a machine does not halt, the claim is about every one of its infinitely many steps, and there is nothing finite to exhibit. No method, given the machine, certifies non-halting in every case. The title of this note comes from the two corollaries of Turing's Undecidability Theorem, which are stated in Section~\ref{sec:events}. ``Safety is $\Pi_1$'' says that the statement that a program has no exploit is a $\Pi_1$ statement, and that no method can certify it from the cases it has examined. Its negation, the statement that the program has an exploit, is a $\Sigma_1$ statement, and an exploit is its witness. ``Provability is $\Sigma_1$'' says that the statement that a sentence has a proof is a $\Sigma_1$ statement in the same way, and that a proof is its witness. Note that a proof of safety, if it exists, is also a witness to the claim that the statement ``a program has no exploit'' is derivable in some theory. But what it proves is $\Pi_1$.

One might hope that the statistical proposer rescues the verifier from undecidability. A verifier alone cannot decide either safety or provability; but a verifier guided by a good proposer, one trained on every exploit or every proof ever written, might seem able to. However, the following proposition says that it cannot. The proposer can change which candidates arrive first, but it cannot change which questions a verifier can settle. And the set of questions it cannot settle is infinite.

\begin{proposition}[Proposer Independence]
\label{prop:proposer}
Let $R$ be a decidable relation between natural numbers and let $Y = \{a : \exists w\, R(a, w)\}$ be the set of $a$ for which some $w$ passes the check. Let a proposer $\pi : \mathbb{N} \times \mathbb{N} \to \mathbb{N}$ and a stopping rule $b : \mathbb{N} \to \mathbb{N}$ be total computable. Let $H(a)$ test $R(a, \pi(a, k))$ for $k = 1, \ldots, b(a)$ in order, answer ``yes'' at the first success, and answer ``no'' if all fail. Let $H_\infty(a)$ do the same without the bound $b$, so that it answers ``yes'' at the first success and otherwise never halts. Then:
\begin{enumerate}
\item[(a)] $H$ is total computable, while $H_\infty$ is partial computable.
\item[(b)] If $H(a) = \text{yes}$ then $a \in Y$, and the $w = \pi(a, k)$ found yields assurance of $a \in Y$, in the sense of Definition~\ref{def:assurance}, in any system of that definition in which $R$ is representable.
\item[(c)] If $Y$ is not decidable, then $H$ answers ``no'' on infinitely many $a \in Y$.
\item[(d)] If $\pi(a, \cdot)$ ranges over all of $\mathbb{N}$ for every $a$, then $H_\infty$ halts with ``yes'' on every $a \in Y$ and never halts on $a \notin Y$.
\end{enumerate}
\end{proposition}

As before, $a$ is a problem, $w$ is a candidate answer, and $R(a, w)$ is the check that $w$ answers $a$. In attacking a program, $a$ is the program $P$, $w$ is an input together with a step count, and $R$ is the relation $\Viol$, which holds when running $P$ on that input for that many steps produces a violation (Section~\ref{sec:security} gives the details). In proving, $a$ is a sentence $\sigma$ to be proved, $w$ is a proof script $p$, and $R$ is $\Prf$. The proposer's two arguments make ``the $k$-th candidate for $a$'' a function: $\pi(a, k)$ is the $k$-th candidate the proposer offers for $a$. The budget $b(a)$ is the number of candidates it is allowed to try on $a$, and it may depend on $a$. The set $Y$ is the set of problems that have an answer, and it depends on $R$: for programs it is the set of unsafe programs, written $\Unsafe$ in Section~\ref{sec:security}, and for proofs it is the set of theorems of $T$. To say that $Y$ is not decidable is to say that no procedure sorts every problem correctly into those that have an answer and those that have none.

Part (c) applies the undecidability of $Y$ to the loop. With a budget, the loop always answers, and by (b) its ``yes'' is always correct. So its errors all lie in its ``no'': there are infinitely many problems that have an answer and on which the loop gives up. It makes no difference whether the proposer or the verifier decides when to give up: a proposer that says ``none of my candidates passed, so there is no answer'' is the loop with a budget. Where a computable bound on the search is known, the loop's ``no'' is correct. But then the problems in question form a decidable class, and part (c) does not apply to them. Without a budget, the loop does not always answer. By (d), it runs forever on a problem that has no answer. The proof is in Appendix~\ref{app:proofs}.

Any proposer can be made to satisfy the range hypothesis of (d) by interleaving it with the enumeration of $\mathbb{N}$. This changes the sequence of candidates and leaves $R$ and $Y$ as they were, so (a) to (c) hold for the new proposer as they did for the old. What a proposer contributes, then, is the order in which candidates arrive. But the order is what is provided by statistics, and (c) shows that no order and no budget turns $H$ into a decider. This is what ``proposer independence'' means. 

Traditional coding and proving is subject to undecidability, and one might have hoped that the statistical step of modern coding and proving removes that limit. It does not. What a better proposer does is to find, within a given budget, the answers to more of the problems that have an answer. It does not make the loop's ``no'' trustworthy: by (c), infinitely many of those answers remain wrong. The proof of (c) uses nothing about $\pi$ except that it is total computable. So (c) holds alike for a proposer that simply lists all candidates in order and for a model trained on every proof ever written. In the hybrid framework's terms, the statistical proposer fixes the order in which candidates arrive, and the deductive verifier assures each candidate that passes. However good the proposer is, it does not change which problems can be settled by the verifier. The reason is simple: when a problem has no answer, every candidate fails, and a sequence of failed candidates looks the same whether no answer exists or the answer has not yet arrived.

\section{The two events under the hybrid framework}
\label{sec:events}

\subsection{Security: safety is \texorpdfstring{$\Pi_1$}{Pi1}}
\label{sec:security}

In the verification literature, a \emph{safety property} is one which says that a bad thing never happens \citep{lamport1977,alpern1985}. Its defining feature, in the words of \citet{alpern1985}, is that if the bad thing happens then there is an identifiable point at which it happens, and that it is then irremediable. A safety property is therefore refuted by a finite prefix of a run and by nothing less. The statement ``The sandbox is never escaped'' expresses one such property. 

Let us now fix a specification $\varphi$ of this kind. Let $\Viol(P, x, s)$ be a decidable relation that holds when program $P$, run on input $x$ for $s$ steps, has produced a violation of $\varphi$. The step bound is what makes the relation decidable. Like $\Test$ in Section~\ref{sec:framework}, $\Viol$ is a relation about a bounded run of a program, and it is checked in the same way, by running the program for $s$ steps. Let us write
\[
\Unsafe = \{P : \exists x\, \exists s\, \Viol(P, x, s)\}.
\]
$\Unsafe$ is $\Sigma_1$, so the safety of $P$ is the $\Pi_1$ statement $\forall x\, \forall s\, \neg\Viol(P, x, s)$.

Safety is a statement about every input and every number of steps. By Definition~\ref{def:assurance}, it is assured only when a derivation of it is available, and testing does not supply one. A test is one instance, $\neg\Viol(P, x_k, s_k)$, and a test harness is a verifier of instances: it compiles the candidate, runs it against the target, and reports. Each passed test is assured, but no finite list of passed tests covers every input. Hence the list does not assure the statement about all of them. Accordingly, a system that has passed every test run on it and holds no derivation of its safety statement stands where the lucky guesser of K\&K stands. There is surely a difference between them: the system is assured of each test it has passed, while the guesser of K\&K is assured of none of the values it has produced, for in the latter case there is no verifier that settles whether a value is the true probability (Section~\ref{sec:framework}). Still, about the next input the system is assured of nothing, just as the lucky guesser of K\&K is. OpenAI's own description of the July sandbox is a description of this position: it had been tested and validated, but the agents chained vulnerabilities that no test had covered.

The attacker's side is the opposite. The statement wanted by an attacker is $P \in \Unsafe$, which is $\Sigma_1$ and assured by just one exploit. An attack agent is the loop of Proposition~\ref{prop:proposer} without a budget. It proposes candidate inputs, runs them, and keeps the one that works. The July agents did this. Undecidability concerns a procedure applied across programs, such as a test suite or a stopping rule, so the corollary is stated for the set of unsafe programs.

\begin{corollary}
\label{cor:unsafe}
Let us suppose that there is an unsafe program $Q$. Let us also suppose that, when a simulation of any Turing machine is run before $Q$ is started, it does not cause a violation while running but does not prevent the violation caused by $Q$ once the simulation has halted. Then $\Unsafe$, which is $\Sigma_1$, is not decidable.
\end{corollary}

The proof, a reduction from the halting set, is in Appendix~\ref{app:proofs}. (It is the reduction used by textbook proofs of Rice's Theorem. But we do not invoke Rice's Theorem itself, which concerns properties of the function computed by a program, while escape is a property of the run.) The two suppositions are what the reduction needs. The first excludes a specification that is not violated by any program, for then $\Unsafe$ is empty and thus trivially decidable. The second says that a violation of $\varphi$ is a property of the run and does not depend on what the program computed before: a computation that is run before $Q$ produces no violation while it runs; once it halts, however, the violation caused by $Q$ still occurs. Without this supposition, $\Viol$ could be any decidable relation on programs, and nothing about $\Unsafe$ would follow from the undecidability of halting. 

Proposition~\ref{prop:proposer}(c) now applies with $Y = \Unsafe$: every computable stopping rule, applied to every proposer of inputs, declares ``no violation found'' on infinitely many unsafe programs. So no stopping rule turns an attacker's silence into assurance of safety. The practical form of this was stated by \citet{dijkstra1970}: testing shows the presence of bugs but never their absence. Our hybrid framework supplies the name for what testing leaves one with: it leaves one a \textit{lucky guesser} at best.

One may think that Turing's Undecidability Theorem can still be avoided in two ways. But neither works. The first is the sandbox itself. A sandbox does not try to show that the program inside it is safe. It is a monitor: it watches each step of the program and blocks the steps forbidden by a policy \citep{schneider2000}. So it may seem that no statement about every input of the program needs to be assured. However, note that a monitor blocks only what passes through it. It keeps the program in only if every path by which the program could get out passes through the monitor. Where the paths are finitely many and can be listed, this can be checked outright, and the risk then lies in whether the list matches the deployed system. Where they cannot be listed, however, it is a statement about every path, and it is $\Pi_1$ in the same way. In July, the registry was a path that was not mediated by the monitor. So a sandbox does not remove the $\Pi_1$ statement that has to be assured. It replaces a statement about the program with a statement about the monitor.

The second is finiteness. Unlike an idealized machine such as a Turing machine, whose memory is unbounded, a \textit{non-ideal} or \emph{real} machine has only a bounded memory, and its state at any moment is the content of that memory. So a machine with $n$ bits of memory has at most $2^n$ states, and its safety can in principle be decided by checking every state it can reach. Such a check would be a proof and not a test, because the states are finitely many and can therefore all be listed, so that the list is \emph{known to be complete}. Strictly, then, Turing's Undecidability Theorem does not apply to a real machine.

However, note that no one can \textit{actually} perform such a check. A real machine has billions of bits of memory, and $2^n$ is then far beyond what any defender can compute. So the defender can test only a small part of the states, and untested states always remain, as they do in the infinite case. For the defender, therefore, they are practically no different from infinitely many, and this is why this note treats the safety of a real system as a $\Pi_1$ statement. What matters in a \textit{real} situation, then, is not whether the instances are finitely or infinitely many. It is whether the statement is settled by a single instance, as a $\Sigma_1$ statement is, or only by something that covers all instances, as a $\Pi_1$ statement is. A list of checked instances, which is finite, covers all of them only if it is known to be complete, and in practice the defender's list of checked states is never complete. So the only way to cover the untested states is a proof that does not go through them one by one. Section~\ref{sec:residual} describes how such a proof works.

Such a proof can be supplied by a compiler, which does check a program against a specification, but only when the specification is written as a type or comes with a proof that is met by the program. ``No escape'' was neither. In July, therefore, there was no such check to run. Instead, there was only a monitor and a record of tests.

\subsection{Mathematics: provability is \texorpdfstring{$\Sigma_1$}{Sigma1}}
\label{sec:proofs}

Let us fix a consistent, recursively axiomatized theory $T$ that extends Robinson arithmetic. Let us write $\Prf(p, \sigma)$ for the primitive recursive relation ``$p$ is a proof of $\sigma$ in $T$'', and $\Prv(\sigma)$ for $\exists p\,\Prf(p, \sigma)$.

\begin{corollary}[Church 1936; Rosser 1936; Tarski, Mostowski and Robinson 1953]
\label{cor:church}
$\Prv$ is $\Sigma_1$ and not decidable.
\end{corollary}

\citet{church1936} proved this under $\omega$-consistency for systems of the kind G\"odel had treated, and \citet{rosser1936} removed $\omega$-consistency in favor of consistency. Robinson arithmetic was introduced later, and the form stated, for every consistent recursively axiomatized extension of it, is \citet{tmr1953}. We should note that, historically, Corollary~\ref{cor:church} is not a corollary of anything. Church's paper appeared before Turing's, and its argument does not go through the halting set. Logically, however, it is Turing's Undecidability Theorem carried into provability, which is why we present it as a corollary here. The halting set reduces to $\Prv$, since every recursively enumerable set $A$ is weakly representable in $T$. That is, there is a formula $\varphi$ such that, for every $n$, $n \in A$ if and only if $T \vdash \varphi(\bar n)$ \citep{ehrenfeucht1960}. The map $n \mapsto \ulcorner \varphi(\bar n) \urcorner$ then reduces $A$ to $\Prv$, and a decision procedure for $\Prv$ would decide $\Halt$.

In the hybrid framework, the statement wanted by a prover is $\Prv(\sigma)$. It is $\Sigma_1$. Its assurance in the sense of Definition~\ref{def:assurance} comes from a witness $p$. The check $\Prf(p, \sigma)$ is a finite computation, and by representability, as in Proposition~\ref{prop:proposer}(b), it yields a derivation of $\exists p\, \Prf(p, \bar\sigma)$ in any system of Definition~\ref{def:assurance} in which $\Prf$ is representable, $T$ itself among them. A theorem is the case in which the whole of assurance is one finite object. The check of a candidate proof is a computation on natural numbers, so it falls under the purely mathematical clause of Definition 3 of K\&K. The October release of manuscripts by OpenAI is this clause applied many times over. For each manuscript with a Lean file, there is a $p$, and the kernel's acceptance of $p$ is the assurance, in Lean's foundation, of the formal statement it proves.

Our hybrid framework also explains what the verifier does not assure. The statement $\neg\Prv(\sigma)$, which means ``$\sigma$ is not provable'', is $\Pi_1$ and thus has no witness. A sequence of failed candidates is a list of instances $\neg\Prf(p_k, \sigma)$, each of them assured, but the statement about all candidates that they are instances of is not assured by any finite list. Still, little is lost by this in mathematics. ``No proof found'' costs only the budget spent on the search. By contrast, in security the same silence costs more than budget. The defender keeps relying on a statement that has not been assured by the silence, and in July the price was the breach itself. Section~\ref{sec:flip} develops this point.

The proposer changes nothing here either. For proofs, take $R = \Prf$, $a = \sigma$, and $w = p$. Then $Y = \Prv$, which is not decidable by Corollary~\ref{cor:church}, and Proposition~\ref{prop:proposer} says that every loop of a computable proposer with a proof verifier and a stopping rule answers ``no'' on infinitely many theorems. Statistics moves theorems into the set reached by a given budget. But it does not move the boundary of $\Prv$, and no run of failures, however long, assures that $\sigma$ has no proof. The run looks the same whether $\sigma$ has no proof or has one that has not yet been reached by the proposer. 

This shows that the division of labor between a statistical proposer and a deductive verifier is forced by what each of them can do, and not chosen by design. The proposer fixes the order in which candidates arrive, but it assures nothing, because it checks nothing. The verifier assures each candidate that passes, but it does not choose which candidates arrive; it can only check those it is given, one at a time. Each does what the other cannot. What neither of them can do is settle that a sentence has no proof, since $\Prv$ is not decidable.

\subsection{Same theorem, opposite desiderata}
\label{sec:flip}

Sections~\ref{sec:security} and \ref{sec:proofs} have the same structure. In both, there is a statistical proposer that orders candidates, a deductive verifier that settles instances, a $\Pi_1$ statement that is not assured by any finite list of instances, and a $\Sigma_1$ statement that is assured by a witness. The hybrid framework is symmetric between them. What is not symmetric is which statement is wanted by each domain.

In security, the wanted statement is the $\Pi_1$ one, $\forall x\, \forall s\, \neg\Viol(P, x, s)$, which says that, for every input $x$ and every number of steps $s$, running the sandbox on $x$ for $s$ steps produces no escape. That is what ``the sandbox is safe'' means, and it is what OpenAI's tests were meant to support. It is not assured by any finite list of instances, and the statement that is assured is its negation, namely the breach $\exists x\, \exists s\, \Viol(P, x, s)$, for which the July agents supplied the witness. The loop's success mode is an assured breach, and its failure mode is silence. Here silence costs more than budget. The unassured statement is the one on which the defender is acting, so silence leaves a system that is relied upon as safe with only a clean record to support it, which is the lucky guesser's position. The cost of that position is paid when the record stops being clean.

In mathematics, the wanted statement is the $\Sigma_1$ one, $\exists p\, \Prf(p, \sigma)$, which says that there is a $p$ that is a proof of $\sigma$, where $\sigma$ is, for example, the zero-free half-plane $\mathrm{Re}(s) > 7/8$ for $\zeta$. It is assured by a specific proof of $\sigma$ such as the Lean file in the October catalog. The unwanted $\Pi_1$ statement $\forall p\, \neg\Prf(p, \sigma)$, which says that no $p$ is a proof of $\sigma$, stays unassured with little cost. This is because, in mathematics, no one takes the loop's silence as evidence that $\sigma$ has no proof, whereas in security a defender does take the clean record as evidence that the sandbox has no exploit. When unprovability matters in mathematics, as it does for the G\"odel sentence of $T$, it is itself assured by a proof in a stronger theory, not by failed searches. In mathematics, therefore, the loop's failure mode costs only budget, while its success mode is assured in full. The statistical proposer may be as wrong as it likes, because the deductive verifier settles each case.

So the two events fall on the same side of one theorem, while the two wanted statements fall on opposite sides. The theorem is Turing's Undecidability Theorem, Corollary~\ref{cor:unsafe} being its form for the July event and Corollary~\ref{cor:church} being its form for the October event. In July, agents found witnesses for a $\Sigma_1$ statement no one wanted true. In October, agents found witnesses for $\Sigma_1$ statements everyone wanted true. What we call \emph{risk} is a situation in which what can be assured by a verifier, an instance, and what is wanted, a statement about all instances, come apart. Turing's Undecidability Theorem shows that no computable method can close the gap in general.

\section{Where the risk goes}
\label{sec:residual}

Section~\ref{sec:security} shows that \emph{testing} cannot assure safety, and a reader will ask what can. The answer is a \emph{proof}. Let $P$ be the particular program in question, for example the July sandbox, and let us write
\[
\mathsf{Safe}(P) := \forall x\, \forall s\, \neg\Viol(P, x, s)
\]
for the $\Pi_1$ statement that $P$ is safe. What can assure the safety of $P$ is a proof of $\mathsf{Safe}(P)$ in the theory $T$ of Section~\ref{sec:proofs}. The same $T$ serves here because $\mathsf{Safe}(P)$ is a statement of arithmetic. Turing's Undecidability Theorem may seem to rule out such a proof, since it shows through Corollary~\ref{cor:unsafe} that there is no decision procedure for safety. But that is a misreading of what undecidability is about. Corollary~\ref{cor:unsafe} says that no single procedure sorts \emph{all} programs into safe and unsafe. It does not say that no \emph{particular} program $P$ can be proved safe, just as the undecidability of $\Prv$ does not say that no particular theorem $\sigma$ can be proved.

A reader may also ask how a finite proof can cover infinitely many inputs. It does not go through them one by one. The usual shape of a proof of $\mathsf{Safe}(P)$ is an invariant, that is, a property of the states of $P$ that (i) holds in every initial state, (ii) is not destroyed by any step of $P$, and (iii) excludes a violation. Each of (i) to (iii) is about a single state or a single step, so it can be established from the code of $P$ without following any run. Together they give $\mathsf{Safe}(P)$ for every input and every number of steps, by induction on the number of steps. This is what Section~\ref{sec:framework} said: a finite derivation covers every run because a universal quantifier does, where no list of runs can. It is what verified software does \citep{klein2009,leroy2009}, and it is the direction taken by proposals for guaranteed safe AI \citep{dalrymple2024}. 

Finding such a proof is itself a search for a witness, namely a proof $p$ with $\Prf(p, \ulcorner \mathsf{Safe}(P) \urcorner)$. So the search is on the $\Sigma_1$ side of Turing's Undecidability Theorem, and Proposition~\ref{prop:proposer} applies to it as to any other search. It may fail, and no rule says when to stop. But when it succeeds, what it assures is the $\Pi_1$ statement $\mathsf{Safe}(P)$, and that is the only route to assurance of safety in the sense of Definition~\ref{def:assurance}. That route is not open for every program: it is closed for an unsafe program, since an exploit exists and will be found by whichever loop reaches it first, the defender's tests or the attacker's search. In July, the attacker's search reached it first. It is closed for some safe programs, too, since some true $\Pi_1$ statements have no proof in $T$. What follows for engineering is that a system that contains an agent should be built so that its safety has a proof before it is deployed, as verified software is built, and not built first and proved afterward. The rest of this section will assume that a proof of safety has been established, and asks where the risk then goes.

A proof of safety accepted by a verifier guarantees that the deployed system is safe only if the following four conditions hold: First, the verifier must be sound; otherwise the proof accepted by it may not be a proof in $T$. Second, $T$ must be consistent; otherwise a statement proved in $T$ may be false. Third, the mathematical definitions $\Dmath$ must be adequate; otherwise $\mathsf{Safe}(P)$ may be true without saying what is meant by safety. Fourth, the schema $\Dsch$ must be adequate; otherwise $P$ may be safe while the deployed system is not. The proof establishes none of these four conditions but rather depends on them. Accordingly, the risk moves to these four places. The first two, namely ``the verifier is sound'' and ``$T$ is consistent'', are again $\Pi_1$ statements, while the last two, about the adequacy of the mathematical definitions and the schema, are questions of another kind. The two events of July and October show each of the four.

Any proof of safety in security depends on these four conditions, whatever the program. The first three are required for any proof in mathematics as well. In both domains, the goal is truth and the assurance is a derivation. In mathematics, the accepted route to the truth of $\sigma$ is a proof in a trusted theory, and the soundness of the verifier, the soundness of that theory (which amounts to its consistency for a $\Pi_1$ theorem), and the adequacy of its definitions are all carried as a shared background that is not re-examined for each theorem. In security, the same three are present, but the adequacy of the schema $\Dsch$ is also required because $\Dsch$ ties the model that was proved safe to the system that was deployed. The adequacy of the schema cannot be treated as background, because an adversary tests it on the deployed system.

\paragraph{The verifier.} A certificate shows that a derivation in $T$ exists only if it is issued by a verifier that accepts only those derivations in $T$. A verifier with this property is called \emph{sound}, and ``the verifier is sound'' is a $\Pi_1$ statement about a program because the verifier is itself a program. The soundness of a verifier must be kept apart from the soundness of the theory $T$, the latter being the property that $T$ proves only true statements. Neither implies the other. A verifier can be sound while the theory implemented by it proves a false statement, and a theory can be sound while its implementation, the verifier, accepts a non-derivation. 

The latter is what actually happened in Lean: Lean 4.32.1 fixed a soundness bug in the kernel that was found with the help of an OpenAI model, and Lean 4.34.0 fixed three soundness issues found by AI adversarial testing and reported by OpenAI, one of which an OpenAI agent turned into two distinct exploits deriving \texttt{False} \citep{lean4321,lean4340}. The October catalog is built on the toolchain released after those fixes, and the statement that the current kernel accepts nothing but derivations in the implemented theory is a $\Pi_1$ statement for which there is no certificate. This $\Pi_1$ statement is about the current kernel, which is a program, so it is of the same kind as safety itself and is handled in the same ways: by checking the certificate again with an independent verifier, by proving the kernel correct in a further theory, or by keeping the kernel small enough to be checked by humans. In an evaluation harness, the \emph{grader} plays the verifier's part, and in July the agents attacked the grader before they attacked anything else. A sandbox has no verifier, since it does not try to show that the program inside it is safe (Section~\ref{sec:security}). What it relies on is the monitor, and the July monitor had a path it did not mediate.

\paragraph{The theory $\boldsymbol{T}$.} Suppose that the verifier is sound. Then a proof of safety shows that the statement $\mathsf{Safe}(P)$ is provable in $T$. But it does not yet show that the statement is \emph{true}, which in general requires the soundness of $T$. Note that, for a $\Pi_1$ theorem such as $\mathsf{Safe}(P)$, its truth requires only the consistency of $T$, which is a weaker property than soundness. To see why, suppose that $\mathsf{Safe}(P)$ is provable in $T$ but false. Since it is false, it has a counterexample, namely numbers $x_0$ and $s_0$ such that $\Viol(P, x_0, s_0)$ is true. The relation $\Viol$ is decidable, and every true instance of a decidable relation is provable in $T$ because $T$ extends Robinson arithmetic and every true $\Sigma_1$ sentence is provable in the latter. So $\Viol(P, x_0, s_0)$ is provable in $T$, which implies that the negation of $\mathsf{Safe}(P)$ is provable in it. Hence both the statement $\mathsf{Safe}(P)$ and its negation are provable in $T$, which means that $T$ is inconsistent.

So whether or not a safety statement proved in $T$ is true depends on the consistency of $T$. But ``$T$ is consistent'' is itself a $\Pi_1$ statement, and G\"odel's Second Incompleteness Theorem implies that no consistent theory that contains enough arithmetic can prove its own consistency. This limitation is unavoidable for us because any theory in which safety proofs are done, including Lean's type theory, does contain enough arithmetic. The consistency of $T$ can be proved by a stronger theory, but then the consistency of the latter must be proved by a still stronger theory, and the regress does not end. So a proof of safety does not remove the dependence on an unproved $\Pi_1$ statement. It does replace one such statement with a \emph{better} one: for each new program $P$, $\mathsf{Safe}(P)$ is a new $\Pi_1$ statement, whereas ``$T$ is consistent'' is a single statement shared by every proof ever done in $T$ and studied on its own for a century. A Lean certificate depends on the same kind of condition, that is, on the consistency of Lean's type theory when the theorem is $\Pi_1$, and on its soundness otherwise.

\paragraph{The mathematical definitions.} The definitions in question are the mathematical definitions $\Dmath$ of the system: in Lean, the definitions of Mathlib, such as that of $\zeta$; in security, the specification $\varphi$, that is, the definition of $\Viol$. A Lean certificate shows that the statement in the file is a theorem. But it does not show that the statement in the file is the one the mathematician had in mind. Whether Mathlib's $\zeta$ and the half-plane in the file are the number theorist's is settled by people who read the definitions, and the October catalog records its own review status as unchecked. A safety proof has the same gap. It assures that the program $P$ meets a formal specification $\varphi$, but it does not guarantee that $\varphi$ really means ``no escape''.

\paragraph{The schema.} The fourth place is a gap that can be found in the domain of security but not in mathematics. Up to this point, we have treated a program as a purely mathematical object. The relation $\Viol$ is defined on a model of execution, $\mathsf{Safe}(P)$ is a statement of arithmetic, and a proof of $\mathsf{Safe}(P)$ is a proof about the model. But the July escape did not happen in a model. It happened on servers. A proof about the model says something about the deployed system only if the deployed system can do nothing that cannot be done by the model, and that is a claim about the empirical world and not a statement of arithmetic. This is where the empirical definitions $\Demp$ of K\&K enter. 

As is explained in K\&K, $\Demp$ has two parts: a schema $\Dsch$ and a record of facts $\Dfact$. The part that matters here is the schema. The record plays no part, because the proof of $\mathsf{Safe}(P)$ uses no recorded fact about the system, and a safety argument that did rely on such facts would be an argument from tests, which has already been treated in Section~\ref{sec:security}. The schema fixes which fact of the world is represented by each symbol of the model. For the sandbox, it fixes which channels of the deployed system count as the inputs $x$ of $\Viol(P, x, s)$ and which events count as a violation. This must be laid down explicitly, because mathematics alone does not say anything about it. The schema is not a premise of the proof of $\mathsf{Safe}(P)$, which remains a purely mathematical proof in $T$. It is what connects the proved statement to the deployed system. 

Inside the system, the schema is a definition, so no derivation can show that it is wrong. Whether it fits the deployed system can be judged only from outside, by comparing the two, whoever does the comparing. In July, the tests that validated the sandbox were tests of a model in which the registry was a channel for fetching packages, and the registry that was deployed could also be made to issue server-side requests. Had there been a proof of $\mathsf{Safe}(P)$ for that model, it would have been valid, and the escape would still have happened. A proof about $\zeta$ has no such gap, because $\zeta$ is not deployed anywhere. This is where security and mathematics diverge, and our hybrid framework locates the divergence in one component, the schema $\Dsch$.

\medskip

To sum up, a proof of $\mathsf{Safe}(P)$ does not make the deployed system safe in any absolute sense. It replaces a $\Pi_1$ statement, which is new for every new system and supported only by instances, with the following four places: the soundness of the verifier, the consistency of the theory $T$, the adequacy of the mathematical definitions, and the adequacy of the schema. All but the schema can be shared across proofs, and the schema is tested on each deployed system. That replacement is the whole of what verification offers, and that is a great deal. What it requires is that one should know where the residual risk went, because that is where the next failure comes from. In July, it was in the schema. In October, it lies in the mathematical definitions and, until the next bug, in the verifier.

\section{Conclusion}
\label{sec:conclusion}

Safety is $\Pi_1$ and provability is $\Sigma_1$. In security, the wanted statement is the $\Pi_1$ statement that no breach exists. What the deductive verifier assures from candidates is a breach. It can also assure the absence of a breach, but only through a proof of the $\Pi_1$ statement, and not for every system. In mathematics, the wanted statement is the $\Sigma_1$ statement that a proof exists, and what the deductive verifier assures from candidates is a proof. Turing's Undecidability Theorem shows that in neither domain does any computable method turn assured instances into an assured statement about all instances, and Proposition~\ref{prop:proposer} shows that no statistical proposer changes this. Assurance holds only when a derivation is available. For a $\Sigma_1$ statement, the derivation is a witness, and for a $\Pi_1$ statement, it is a proof in a theory, after which the residual risk sits in the verifier, the theory $T$, the mathematical definitions, and the schema.

This note is based on known results: Turing's Undecidability Theorem, the reduction from the halting set behind Corollary~\ref{cor:unsafe}, Church's result behind Corollary~\ref{cor:church}, the notion of a safety property \citep{lamport1977,alpern1985}, Dijkstra's remark on testing \citep{dijkstra1970}, the exploit as a witness \citep{clarke1999,avgerinos2011}, the arithmetical hierarchy as a classification of what can be verified and what can be refuted \citep{kelly1996}, and the undecidability of containment \citep{alfonseca2021}. Our hybrid framework puts them under a single definition of assurance. Under it, the agents that breached Hugging Face and the agents that produced Lean proofs of a zero-free region for $\zeta$ were doing the same kind of thing, although the two domains differ in the statement each one wants. We hope that this analysis is of use for what agentic AI faces next: proofs of safety for the systems that contain agents, and a clear view of where the risk remains once such a proof has been obtained.

\paragraph{Acknowledgments.} Jinho Kang was supported by Gwangju Institute of Science and Technology. Jinsook Kim was supported by Korea Institute for Advanced Study. 

\paragraph{AI Use Disclosure.} We wrote and verified this note with the assistance of AI tools. We take full responsibility for its content, and the errors are ours.

\appendix
\section{Proofs}
\label{app:proofs}

\paragraph{Proof of Proposition~\ref{prop:proposer}.}
(a) $b(a)$ is computed in finitely many steps since $b$ is total. For each $k \le b(a)$, $\pi(a, k)$ is computed in finitely many steps since $\pi$ is total, and $R(a, \pi(a, k))$ is decided in finitely many steps since $R$ is decidable. There are at most $b(a)$ tests, so $H$ halts. $H_\infty$ runs the same procedure without the bound, so it is computable, and it halts on $a$ exactly when some $\pi(a, k)$ passes the check. Hence it is partial computable.

(b) ``Yes'' is answered only when $R(a, \pi(a, k))$ holds for some $k$. By the definition of $Y$, $a \in Y$. Now let $\mathcal{S}$ be a system as in Definition~\ref{def:assurance} and let $\rho$ represent $R$ in $\mathcal{S}$. Since $R(a, w)$ holds, $\mathcal{S} \vdash \rho(\bar a, \bar w)$ by representability, and hence $\mathcal{S} \vdash \exists w\, \rho(\bar a, w)$ by existential generalization. The last sentence expresses $a \in Y$. So the $w$ found makes a derivation of $a \in Y$ in $\mathcal{S}$ available, and that is assurance of $a \in Y$ by Definition~\ref{def:assurance}. 

(c) Suppose $M = \{a \in Y : H(a) = \text{no}\}$ is finite. Define $H'(a)$ to be ``yes'' if $a \in M$ and $H(a)$ otherwise. $M$ is finite, so membership in $M$ is decided by a table, and $H'$ is total computable by (a). If $a \in Y$ and $a \in M$ then $H'(a) = \text{yes}$. If $a \in Y$ and $a \notin M$ then $H(a) \ne \text{no}$ by the definition of $M$, and $H$ has only two outputs, so $H(a) = \text{yes}$ and $H'(a) = \text{yes}$. Conversely, if $H'(a) = \text{yes}$ then either $a \in M \subseteq Y$ or $H(a) = \text{yes}$, and then $a \in Y$ by (b). So $H'$ decides $Y$, against the hypothesis.

(d) If $a \in Y$ there is a $w$ with $R(a, w)$, and $w = \pi(a, k)$ for some $k$ by the range hypothesis. The first such $k$ is reached in finitely many steps and $H_\infty$ halts with ``yes''. If $a \notin Y$ no test succeeds and $H_\infty$ runs forever. \qed

\paragraph{Proof of Corollary~\ref{cor:unsafe}.}
For each index $e$ let $P_e$ be the program that, on input $x$, first simulates machine $e$ on the empty tape and, if the simulation halts, then behaves as $Q$ on $x$. The map $e \mapsto P_e$ is computable. If $e \in \Halt$ the simulation halts, and $P_e$ then violates $\varphi$ on the input on which $Q$ does, after a larger number of steps, by the second part of the second hypothesis, so $P_e \in \Unsafe$. If $e \notin \Halt$ the simulation never halts, and by the first part of the second hypothesis no violation occurs, so $P_e \notin \Unsafe$. Hence $\Halt$ is many-one reducible to $\Unsafe$. If $\Unsafe$ were decidable, then $e \mapsto P_e$ followed by that decision procedure would decide $\Halt$, against Turing's Undecidability Theorem. So $\Unsafe$ is not decidable. That $\Unsafe$ is $\Sigma_1$ is immediate from its definition and the decidability of $\Viol$. \qed

\end{document}